\documentclass[final]{cvpr}

\usepackage{times}
\usepackage{epsfig}
\usepackage{graphicx}
\usepackage{amsmath}
\usepackage{amssymb}

\usepackage{multirow}

\usepackage[pagebackref=true,breaklinks=true,colorlinks,bookmarks=false]{hyperref}

\def\cvprPaperID{28} 
\def\confYear{CVPR CLIC 2021}

\begin{document}

\title{Learning-based Compression for Material and Texture Recognition}

\author{Yingpeng Deng\\
Arizona State University\\
{\tt\small ypdeng@asu.edu}
\and
Lina J. Karam\\
Lebanese American University\\
{\tt\small karam@asu.edu}
}

\maketitle

\begin{abstract}
  Learning-based image compression was shown to achieve a competitive performance with state-of-the-art transform-based codecs. This motivated the development of new learning-based visual compression standards such as JPEG-AI. Of particular interest to these emerging standards is the development of learning-based image compression systems targeting both humans and machines. This paper is concerned with learning-based compression schemes whose compressed-domain representations can be utilized to perform visual processing and computer vision tasks directly in the compressed domain. Such a characteristic has been incorporated as part of the scope and requirements of the new emerging JPEG-AI standard. In our work, we adopt the learning-based JPEG-AI framework for performing material and texture recognition using the compressed-domain latent representation at varing bit-rates. For comparison, performance results are presented using compressed but fully decoded images in the pixel domain as well as original uncompressed images. The obtained performance results show that even though decoded images can degrade the classification performance of the model trained with original images, retraining the model with decoded images will largely reduce the performance gap for the adopted texture dataset. It is also shown that the compressed-domain classification can yield a competitive performance in terms of Top-1 and Top-5 accuracy while using a smaller reduced-complexity classification model.
\end{abstract}

\section{Introduction}
\label{sec:intro}

Learning-based image coding algorithms have shown a competitive compression efficiency as compared to the conventional compression methods, by using advanced deep learning techniques. Specifically, when compared to JPEG/JPEG 2000 standard, learning-based codecs can produce higher compression quality in terms of some common perceptual objective quality metrics such as PSNR (peak signal-to-noise ratio) and SSIM (structural similarity)~\cite{wang2004image} / MS-SSIM (multi-scale structural similarity)~\cite{wang2003multiscale} for some target compression bitrates~\cite{balle2018variational, Theis2017a}.

\begin{figure}[t]
	\centering
	\includegraphics[width=0.3\textwidth]{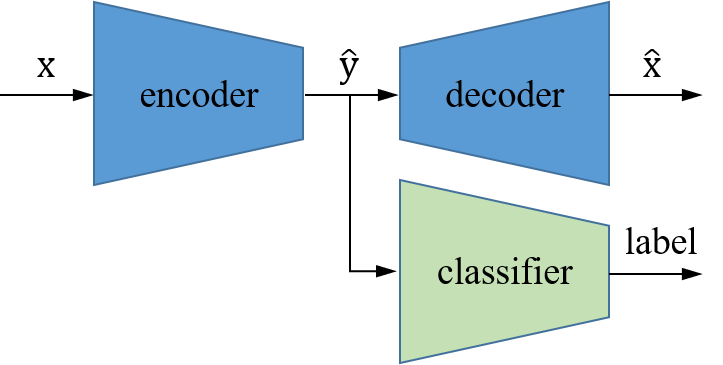}
	\caption{Block diagram of compressed-domain classification.}
	\label{fig:flow1}
\end{figure}

Furthermore, without fully decoding the images, learning-based codecs can be adapted to support image processing and computer vision tasks in the compressed-domain. This latter feature was incorporated as part of the emerging JPEG-AI standard~\cite{jpegai_use}. Figure~\ref{fig:flow1} shows an example for a compressed-domain classification task. Given an original input image $x$, a learning-based codec can encode it and quantize it to form a compressed-domain latent representation $\hat{y}$ (compressed representation). This compressed representation $\hat{y}$ can be losslessly transformed into a bitstream via entropy encoding and then recovered back to the compressed-domain through an entropy decoding process. The decoded image $\hat{x}$ needs to be reconstructed by inputing the compressed representation $\hat{y}$ to the learning-based decoder, which can be abandoned when targeting a machine given that the compressed representation $\hat{y}$ may contain useful information to perform image processing and computer vision tasks in the compressed domain~\cite{jpegai_eval}.

One application of interest is compressed-domain classification. Recently, Torfason~\textit{et al.}~\cite{torfason2018towards} demonstrated that two computer vision tasks, image classification and semantic segmentation, can achieve comparable performance using compressed-domain representations as compared to the tasks performed using pixel-domain decompressed images. In this work, we adopt the JPEG-AI framework~\cite{jpegai_eval} for compressed-domain material and texture recognition. We first evaluate the effects of decoded images on recognition model performance. We then present a system for integrating compressed-domain material and texture recognition, an advanced learning-based image coding network~\cite{balle2018variational}. The rest of our paper is organized as follows. Related works about texture recognition and learning-based image coding algorithms are briefly discussed in Section~\ref{sec:rw}. Section~\ref{sec:alg} describes the compressed-domain material and texture classification method. Performance results and comparison with pixel-domain classification are presented in Section~\ref{sec:exp}, followed by a conclusion in Section~\ref{sec:con}.

\section{Related Works}
\label{sec:rw}

Many effective learning-based image coding architecture are based deep neural networks (DNNs). Theis \textit{et al.}~\cite{Theis2017a} defined a compressive autoencoder to learn the end-to-end trained encoder-decoder network for compression, with a joint rate-distortion loss function. With similar thoughts, Ball\'{e} \textit{et al.}~\cite{balle2016end} described a DNN-based image compression architecture consisting of linear convolutional layers and nonlinear activation functions, and developed an end-to-end optimization framework via backpropagation for rate-distortion performance trade-off. Further improving the prior work, Ball\'{e} \textit{et al.}~\cite{balle2018variational} introduced a hyperprior network to learn and compress the spatial dependency information in the compressed-domain and adopted the MS-SSIM index into the objective function for better visual quality.

Recent texture recognition works turned to adaptive modifications of end-to-end DNN architectures such as the ResNet architecture~\cite{he2016deep}. By adopting ResNet as backbone, Zhang \textit{et al.}~\cite{zhang2017deep} proposed a deep texture encoding network by inserting a residual encoding layer with learnable dictionary codewords before the decision layer. To improve on the method of Zhang \textit{et al.}~\cite{zhang2017deep}, Xue \textit{et al.}~\cite{xue2018deep} combined the global average pooling features with the encoding pooling features through a bilinear model~\cite{freeman1997learning} in their deep encoding pooling network, followed by a multi-level texture encoding and representation~\cite{hu2019multi} to aggregate multi-stage features extracted using the DEP module.

\section{Compressed-Domain Material and Texture Recognition}
\label{sec:alg}

In our work, the texture classification task can be directly performed on the compressed-domain latent representation that is produced by a learning-based image codec. We adopt the variational image compression with a scale hyperprior~\cite{balle2018variational} using the MS-SSIM quality metric in the loss function (HyperMS-SSIM). As shown in Figure~\ref{fig:flow1}, instead of using the fully decoded pixel-domain images to train/evaluate the classification model, the compressed-domain representation $\hat{y}$ is directly taken as the input of the classification model by assuming that it already contains useful information for classification. Given an input image of $H \times W \times 3$, the compressed representation will be $\frac{H}{16} \times \frac{W}{16} \times C$.

\begin{table}[t]
	\centering
	\caption{ResNet-50 and cResNet-39 model architectures used for texture classification.}
	\label{tab:arch}
	\small
	\resizebox{0.4\textwidth}{!}{
		\begin{tabular}{|c|c|cc|}
			\hline
			\textbf{Layer}&\textbf{ResNet-50}&\multicolumn{2}{|c|}{\textbf{cResNet-39}}\\
			\hline
			conv1&$\left[\begin{array}{c}3 \times 3, 64\\3 \times 3, 64\\3 \times 3, 128\end{array}\right]$\footnote&\multicolumn{2}{|c|}{None}\\
			\hline
			\multirow{2}*[-3ex]{conv\_2x}&$3 \times 3$ maxpool&\multirow{2}*[-1.5ex]{$\left[\begin{array}{c}1 \times 1, 32\\3 \times 3, 32\\1 \times 1, 128\end{array}\right]$}&\multirow{2}*[-1.5ex]{$\left[\begin{array}{c}1 \times 1, 32\\3 \times 3, 32\\1 \times 1, 128\end{array}\right]$}\\
			\cline{2-2}
			~&$\left[\begin{array}{c}1 \times 1, 64\\3 \times 3, 64\\1 \times 1, 256\end{array}\right] \times 3$&~&~\\
			\hline
			conv\_3x&\multicolumn{3}{|c|}{$\left[\begin{array}{c}1 \times 1, 128\\3 \times 3, 128\\1 \times 1, 512\end{array}\right] \times 4$}\\
			\hline
			conv\_4x&\multicolumn{3}{|c|}{$\left[\begin{array}{c}1 \times 1, 256\\3 \times 3, 256\\1 \times 1, 1024\end{array}\right] \times 6$}\\
			\hline
			conv\_5x&\multicolumn{3}{|c|}{$\left[\begin{array}{c}1 \times 1, 512\\3 \times 3, 512\\1 \times 1, 2048\end{array}\right] \times 3$}\\
			\hline
			&\multicolumn{3}{|c|}{average pool, 23-d fc, softmax}\\
			\hline
			
	\end{tabular}}
\end{table}

\begin{figure*}[!tb]
	\centering
	\begin{minipage}{0.16\textwidth}
		\centering
		original
	\end{minipage}
	\begin{minipage}{0.24\textwidth}
		\centering
		HyperMS-SSIM-8
	\end{minipage}
	\begin{minipage}{0.24\textwidth}
		\centering
		HyperMS-SSIM-4
	\end{minipage}
	\begin{minipage}{0.24\textwidth}
		\centering
		HyperMS-SSIM-1
	\end{minipage}\\
	\begin{minipage}{0.16\textwidth}
		\includegraphics[width=0.98\textwidth]{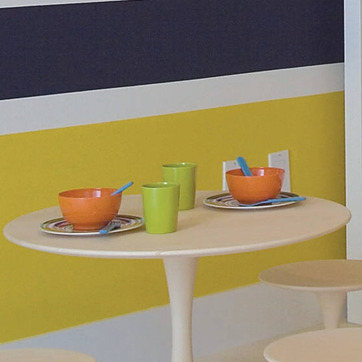}
	\end{minipage}
	\begin{minipage}{0.16\textwidth}
		\flushright
		\includegraphics[width=0.98\textwidth]{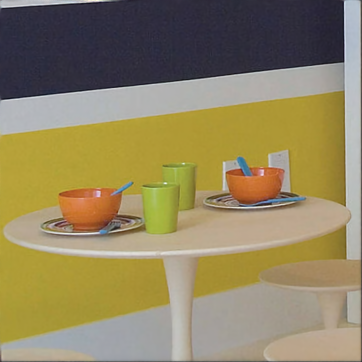}
	\end{minipage}
	\begin{minipage}{0.08\textwidth}
		\flushleft
		\includegraphics[width=0.98\textwidth]{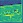}
		\includegraphics[width=0.98\textwidth]{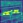}
	\end{minipage}
	\begin{minipage}{0.16\textwidth}
		\flushright
		\includegraphics[width=0.98\textwidth]{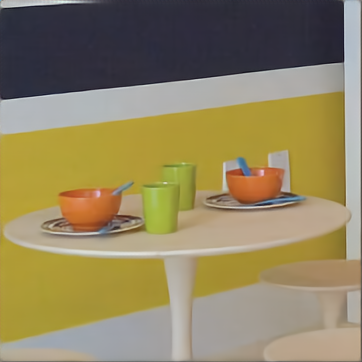}
	\end{minipage}
	\begin{minipage}{0.08\textwidth}
		\flushleft
		\includegraphics[width=0.98\textwidth]{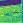}
		\includegraphics[width=0.98\textwidth]{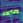}
	\end{minipage}
	\begin{minipage}{0.16\textwidth}
		\flushright
		\includegraphics[width=0.98\textwidth]{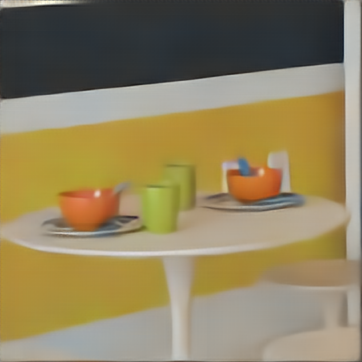}
	\end{minipage}
	\begin{minipage}{0.08\textwidth}
		\flushleft
		\includegraphics[width=0.98\textwidth]{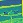}
		\includegraphics[width=0.98\textwidth]{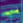}
	\end{minipage}\\
	\begin{minipage}{0.16\textwidth}
		\includegraphics[width=0.47\textwidth]{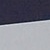}
		\includegraphics[width=0.47\textwidth]{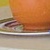}
	\end{minipage}
	\begin{minipage}{0.16\textwidth}
		\flushright
		\includegraphics[width=0.47\textwidth]{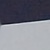}
		\includegraphics[width=0.47\textwidth]{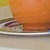}
	\end{minipage}
	\begin{minipage}{0.08\textwidth}
		\ 
	\end{minipage}
	\begin{minipage}{0.16\textwidth}
		\flushright
		\includegraphics[width=0.47\textwidth]{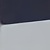}
		\includegraphics[width=0.47\textwidth]{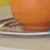}
	\end{minipage}
	\begin{minipage}{0.08\textwidth}
		\ 
	\end{minipage}
	\begin{minipage}{0.16\textwidth}
		\flushright
		\includegraphics[width=0.47\textwidth]{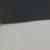}
		\includegraphics[width=0.47\textwidth]{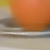}
	\end{minipage}
	\begin{minipage}{0.08\textwidth}
		\ 
	\end{minipage}\\
	\begin{minipage}{0.16\textwidth}
		\ 
	\end{minipage}
	\begin{minipage}{0.16\textwidth}
		\centering
		PSNR: 31.61 dB
		SSIM: 0.9756
	\end{minipage}
	\begin{minipage}{0.08\textwidth}
		\ 
	\end{minipage}
	\begin{minipage}{0.16\textwidth}
		\centering
		PSNR: 30.44 dB
		SSIM: 0.9410
	\end{minipage}
	\begin{minipage}{0.08\textwidth}
		\ 
	\end{minipage}
	\begin{minipage}{0.16\textwidth}
		\centering
		PSNR: 26.80 dB
		SSIM: 0.8883
	\end{minipage}
	\begin{minipage}{0.08\textwidth}
		\ 
	\end{minipage}\\
	\tiny
	\ \\
	\_\_\_\_\_\_\_\_\_\_\_\_\_\_\_\_\_\_\_\_\_\_\_\_\_\_\_\_\_\_\_\_\_\_\_\_\_\_\_\_\_\_\_\_\_\_\_\_\_\_\_\_\_\_\_\_\_\_\_\_\_\_\_\_\_\_\_\_\_\_\_\_\_\_\_\_\_\_\_\_\_\_\_\_\_\_\_\_\_\_\_\_\_\_\_\_\_\_\_\_\_\_\_\_\_\_\_\_\_\_\_\_\_\_\_\_\_\_\_\_\_\_\_\_\_\_\_\_\_\_\_\_\_\_\_\_\_\_\_\_\_\_\_\_\_\_\_\_\_\_\_\_\_\_\_\_\_\_\_\_\_\_\_\_\_\_\_\_\_\_\_\_\_\_\_\_\_\_\_\_\_\_\_\_\_\_\_\_\_\_\_\_\_\_\_\_\_\_\_\_\_\_\_\_\_\_\_\_\_\_\_\_\_\_\_\_\_\_\_\_\_\_\_\_\_\_\_\_\_\_\_\_\_\_\_\_\_\_\_\_\_\_\_\_\_\_\_\_\_\_\_\_\\
	\ \\
	\normalsize
	\begin{minipage}{0.16\textwidth}
		\includegraphics[width=0.98\textwidth]{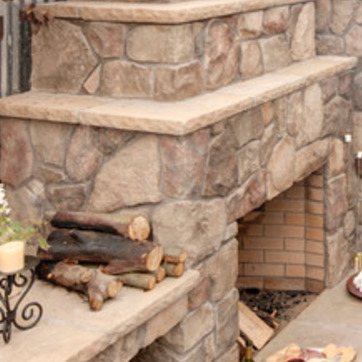}
	\end{minipage}
	\begin{minipage}{0.16\textwidth}
		\flushright
		\includegraphics[width=0.98\textwidth]{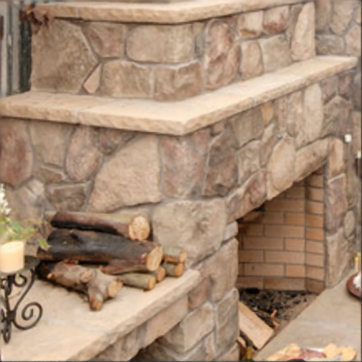}
	\end{minipage}
	\begin{minipage}{0.08\textwidth}
		\flushleft
		\includegraphics[width=0.98\textwidth]{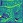}
		\includegraphics[width=0.98\textwidth]{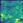}
	\end{minipage}
	\begin{minipage}{0.16\textwidth}
		\flushright
		\includegraphics[width=0.98\textwidth]{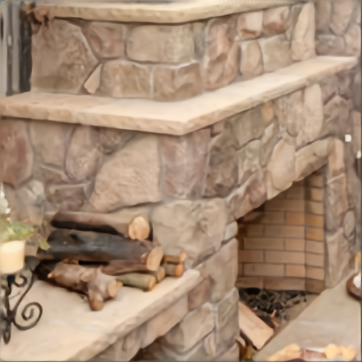}
	\end{minipage}
	\begin{minipage}{0.08\textwidth}
		\flushleft
		\includegraphics[width=0.98\textwidth]{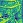}
		\includegraphics[width=0.98\textwidth]{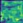}
	\end{minipage}
	\begin{minipage}{0.16\textwidth}
		\flushright
		\includegraphics[width=0.98\textwidth]{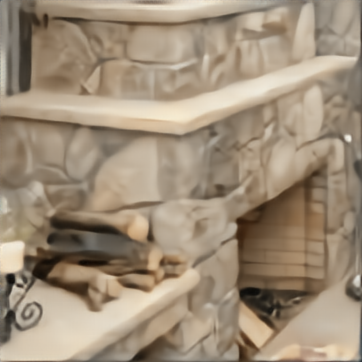}
	\end{minipage}
	\begin{minipage}{0.08\textwidth}
		\flushleft
		\includegraphics[width=0.98\textwidth]{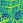}
		\includegraphics[width=0.98\textwidth]{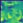}
	\end{minipage}\\
	\begin{minipage}{0.16\textwidth}
		\includegraphics[width=0.47\textwidth]{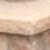}
		\includegraphics[width=0.47\textwidth]{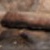}
	\end{minipage}
	\begin{minipage}{0.16\textwidth}
		\flushright
		\includegraphics[width=0.47\textwidth]{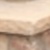}
		\includegraphics[width=0.47\textwidth]{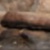}
	\end{minipage}
	\begin{minipage}{0.08\textwidth}
		\ 
	\end{minipage}
	\begin{minipage}{0.16\textwidth}
		\flushright
		\includegraphics[width=0.47\textwidth]{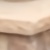}
		\includegraphics[width=0.47\textwidth]{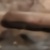}
	\end{minipage}
	\begin{minipage}{0.08\textwidth}
		\ 
	\end{minipage}
	\begin{minipage}{0.16\textwidth}
		\flushright
		\includegraphics[width=0.47\textwidth]{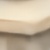}
		\includegraphics[width=0.47\textwidth]{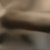}
	\end{minipage}
	\begin{minipage}{0.08\textwidth}
		\ 
	\end{minipage}\\
	\begin{minipage}{0.16\textwidth}
		\ 
	\end{minipage}
	\begin{minipage}{0.16\textwidth}
		\centering
		PSNR: 31.85 dB
		SSIM: 0.9844
	\end{minipage}
	\begin{minipage}{0.08\textwidth}
		\ 
	\end{minipage}
	\begin{minipage}{0.16\textwidth}
		\centering
		PSNR: 30.15 dB
		SSIM: 0.9388
	\end{minipage}
	\begin{minipage}{0.08\textwidth}
		\ 
	\end{minipage}
	\begin{minipage}{0.16\textwidth}
		\centering
		PSNR: 25.09 dB
		SSIM: 0.7920
	\end{minipage}
	\begin{minipage}{0.08\textwidth}
		\ 
	\end{minipage}\\
	\begin{minipage}{0.08\textwidth}
		\tiny \ 
	\end{minipage}
	\caption{Visual examples of the original images (column 1), and decoded images using HyperMS-SSIM-8 (highest quality, column 2), HyperMS-SSIM-4 (middle quality, column 3) and HyperMS-SSIM-1 (lowest quality, column 4) from the class "plastic" (top row) and "stone" (bottom row). Close-ups are shown below each image. At the right of each decoded image are the cumulative compressed-domain representation $\hat{y}$ (top) and standard deviation representation $\hat{\sigma}$ (bottom). PSNR and SSIM values for each decoded image are given.}
	\label{fig:vis}
\end{figure*}

Although many ResNet variants have been proposed for texture recognition, a recent study shows that these modifications based on the ResNet backbone result in trivial improvements or even worse performances~\cite{deng2020study}. So in this paper we only consider ResNet-50. As shown in Table~\ref{tab:arch}, we adopt cResNet-39~\cite{torfason2018towards} as the compressed-domain classification model by removing the first 11 layers of ResNet-50~\cite{he2016deep}. In addition, HyperMS-SSIM learns the standard deviation maps $\hat{\sigma}$ to help the entropy coding process. We find that $\hat{\sigma}$ can also improve the compressed-domain performance while it requires significantly less bits than the compressed-domain representations. Thus, we feed both $\hat{y}$ and $\hat{\sigma}$ into separate residual blocks and concatenate the outputs together as the input of the cResNet-39 model. Table~\ref{tab:arch} gives the details for both ResNet-50 and cResNet-39.

\footnotetext{This layer is different from the original ResNet-50~\cite{he2016deep}, which can improve the texture recognition performance according to~\cite{zhang2017deep}.}

For performance comparison, we also define three different anchors. For all the three anchors, the original input images are encoded and fully decoded, then the classification is performed on the fully decoded images. The three considered anchors are described below:
\begin{itemize}\vspace*{-1.5ex}
	\item \textbf{Anchor 1} - Pixel-domain at decoder with pretrained classifier. The classification task is performed after the decoder, in the uncompressed pixel-domain, by applying a ResNet-50 model that is pretrained using the training images from the considered uncompressed texture dataset.\vspace*{-1.5ex}
	\item \textbf{Anchor 2} – Pixel-domain at decoder with retrained classifier. The classification task is performed after the decoder, in the uncompressed pixel-domain, by applying a ResNet-50 model that is retrained using decoded images from the considered texture dataset.\vspace*{-1.5ex}
	\item \textbf{Anchor 3} - Pixel-domain without further encoding. The classification task is performed directly using the original pixel-domain input image by applying a ResNet-50 model that is pretrained on the original uncompressed texture dataset.
\end{itemize}

\section{Experimental Results}
\label{sec:exp}

Our experiments for texture recognition were performed using the Materials in Context (MINC) Database~\cite{bell15minc}. Specifically, its subset MINC-2500 is adopted. MINC-2500 contains 57,500 images of 23 classes. Each class contains 2,500 images. We use the train-validation-test split 1 provided in the dataset, with 2125 training images, 125 validation images and 250 testing images for each class. We employ the HyperMS-SSIM tensorflow-compression implementation available in~\cite{tfc}, which provides 8 trained models for HyperMS-SSIM corresponding to different quality levels, and thus rates, named as “bmshj2018-hyperprior-msssin-[1-8]” (1 for lowest quality/rate and 8 for highest quality/rate). The compression is done by applying the pretrained “bmshj2018-hyperprior-msssim-1” (HyperMS-SSIM-1), “bmshj2018-hyperprior-msssim-4” (HyperMS-SSIM-4) and “bmshj2018-hyperprior-msssim-8” (HyperMS-SSIM-8) models which correspond to the lowest, middle, and highest quality/rate, respectively, on all images in the MINC-2500 dataset.

\textbf{Anchor training.} The input images are resized to $256 \times 256$ and then randomly cropped to $224 \times 224$, followed by a random horizontal flipping. Standard color augmentation and PCA-based noise are used as in~\cite{zhang2017deep}. All anchor ResNet-50 classifiers are finetuned starting from the ImageNet-trained ResNet-50 model on either the original (Anchors 1\&3) or the decoded (Anchor 2) MINC-2500 training set. The batch size is set to 32. A stochastic gradient descent (SGD) optimizer with momentum of 0.9 and weight decay of 0.0005 is used. The learning rate is initialized to 0.01 and divided by a factor of 10 after 10 and 20 epochs for a total of 30 epochs.

\begin{figure*}[!tb]
	\centering
	\begin{minipage}{0.3\textwidth}
		\includegraphics[width=1\textwidth]{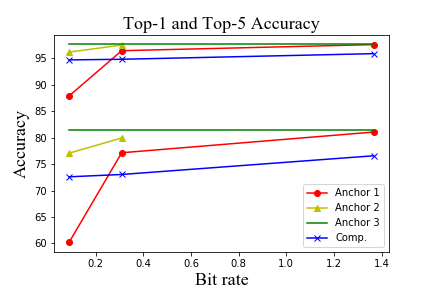}
	\end{minipage}
	\begin{minipage}{0.3\textwidth}
		\includegraphics[width=1\textwidth]{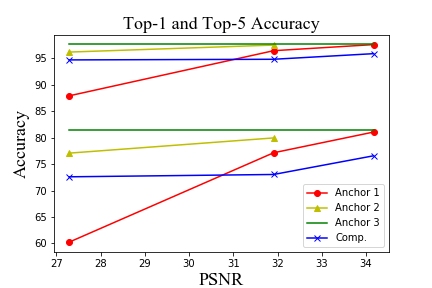}
	\end{minipage}
	\begin{minipage}{0.3\textwidth}
		\includegraphics[width=1\textwidth]{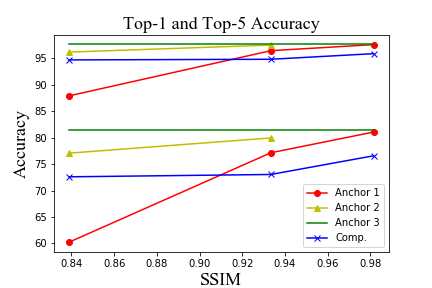}
	\end{minipage}
	
	\caption{Curves for Top-1 (lower part) \& Top-5 (upper part) accuracy results vs. compression rate/qualities (bit rate, PSNR and SSIM).}
	\label{fig:relate}
\end{figure*}

\textbf{Compressed-domain training.} The input compressed-domain representations are resized to $32 \times 32$ and then randomly cropped to $28 \times 28$, followed by a random horizontal flipping. The cResNet-39 classifier is trained using the compressed-domain representations of the MINC-2500 training images starting from the ImageNet-trained cResNet-39 network. The hyperparameter settings are similar to the Anchor training.

\subsection{Classification Performances}

\begin{table}[t]
	\centering
	\caption{Top-1 \& Top-5 accuracy results for the different anchors and the compressed-domain classification at various qualities/rates.}
	\label{tab:acc}
	\resizebox{0.47\textwidth}{!}{
		\begin{tabular}{|c||c|c|c|}
			\hline
			\textbf{Top-1 ACC}&\textbf{HyperMS-SSIM-1}&\textbf{HyperMS-SSIM-4}&\textbf{HyperMS-SSIM-8}\\
			\hline
			Anchor 1&60.26&77.13&81.04\\
			\hline
			Anchor 2&77.06&79.93&-\footnotemark\\
			\hline
			Comp.&72.59&73.03&76.56\\
			\hline
			\hline
			\textbf{Top-5 ACC}&\textbf{HyperMS-SSIM-1}&\textbf{HyperMS-SSIM-4}&\textbf{HyperMS-SSIM-8}\\
			\hline
			Anchor 1&87.88&96.40&97.55\\
			\hline
			Anchor 2&96.14&97.46&-\\
			\hline
			Comp.&94.66&94.78&95.84\\
			\hline
	\end{tabular}}
\end{table}

Here we show the Top-1 and Top-5 accuracy results for the anchors and compressed-domain classification whose images/representations are produced by HyperMS-SSIM-1, HyperMS-SSIM-4 and HyperMS-SSIM-8. For comparison, the resulting Top-1 and Top-5 classification accuracies are 81.50\% and 97.58\%, respectively

\footnotetext{Since the accuracy drops from Anchor 3 to Anchor 1 are trivial for HyperMS-SSIM-8, we don't provide the accuracy results for Anchor 2 which needs retraining.}

From Table~\ref{tab:acc}, it can be seen that, using the highest quality/rate compression model (HyperMS-SSIM-8) results in a negligible drop in classification performance when the decoded images (Anchor 1) are used in place of the original images (Anchor 3) as input to the classifier. At lower quality levels/bitrates (HyperMS-SSIM-1 and HyperMS-SSIM-4), there is a clear decrease in classification performance. Retraining the ResNet50 classifiers using lower quality decoded training images (Anchor 2) results in significant classification performance improvement as compared to using the original pre-trained ResNet50 classifier (Anchor 1). The compressed-domain classification using the finetuned cResNet39 yields a good but lower classification performance in terms of Top-1 and Top-5 accuracy as compared to Anchor 2. Although there still exists some performance gap between compressed-domain and Anchor 2 results, the compressed-domain classification can be more efficient given that it does not need full decoding for the inference process and that it is implemented using a shallower network (cResNet-39), as compared to Anchor 2.

Figure~\ref{fig:vis} shows some visual examples, illustrating the visual quality of decoded texture images that were compressed using HyperMS-SSIM-8, HyperMS-SSIM-4, HyperMS-SSIM-1, in addition to the corresponding original image. For each decoded image, the corresponding cumulative compressed-domain represetation and standard deviation representation, which are obtained by summing up the representation over the channel dimension, are given.

\subsection{Relation to Compression Quality}

Figure~\ref{fig:relate} shows plots of the Top-1 (lower part in each plot) and Top-5 (upper part in each plot) accuracy results versus both objective (bit rate) and subjective (PSNR and SSIM) compression quality metrics for the three adopted learning-based compression models on the MINC-2500 test split. It can be seen that, the Top-1 and Top-5 accuracy results for the same model follow the similar trends. For Anchor 1, the accuracy results seem to increase faster at lower compression quality, while the opposite is observed for compressed-domain classification. By assuming that the performance of Anchor 2 is between those of Anchor 1 and 3, we can observe a roughly linear proportion between Anchor 2 accuracy and PSNR/SSIM.

\section{Conclusion}
\label{sec:con}

Based on the MINC-2500 dataset, we examine the material and texture recognition task with a learning-based compression network HyperMS-SSIM. Specifically, we train/evaluate the ResNet-50 classification model using original/decoded images and the cResNet-39 model using compressed-domain representations. After retraining with the decoded images at middle rate/quality, the ResNet-50 model with decoded images can achieve Top-1 and Top-5 accuracy results that are close to the ones obtained when using original uncompressed images. Furthormore, the retrained ResNet-50 model results in significantly classification performance improvement for images coded at the lowest rate/quality (HyperMS-SSIM-1) as compared to uncompressed images. For compressed-domain classification, the cResNet-39 model provides a good but lower accuracy results yet with no decoding process and less model layers as compared to Anchor 2. Also, we show the relation between classification performance and compression quality by plotting the accuracy curves versus the bit rate and compression quality metrics, PSNR and SSIM.

{\small
\bibliographystyle{ieee_fullname}
\bibliography{egbib}
}

\end{document}